\documentclass[11pt, a4paper, logo, onecolumn]{googledeepmind}

\pdftrailerid{redacted}
\usepackage{bbding}

\usepackage{xspace}
\newcommand{\THFM}{THFM\xspace}

\usepackage{colortbl}
\definecolor{minetable1colorx}{rgb}{0.75, 0.75, 0.75}
\newcommand{\mineyes}{{\scriptsize\CheckmarkBold}}
\newcommand{\mineno}{{\scriptsize \textcolor{minetable1colorx}{\XSolidBrush}}}

\definecolor{cvprblue}{rgb}{0.21,0.49,0.74}
\usepackage[pagebackref,breaklinks,allcolors=cvprblue, colorlinks=true]{hyperref} %
\usepackage{multirow}
\usepackage{multicol}
\usepackage{gensymb}
\usepackage{graphicx}
\usepackage{caption}

\definecolor{figtextcolor}{HTML}{333333}
\definecolor{perceptioncol}{HTML}{1A73E8}
\definecolor{modelingcol}{HTML}{925BCF}
\definecolor{manipulationcol}{HTML}{C440A7}
\definecolor{reasoningcol}{HTML}{DA2F77}

\usepackage{subcaption}
\usepackage{opensans}
\DeclareCaptionFont{subfigcapsize}{\fontsize{8pt}{8pt}\selectfont}  %
\usepackage{xurl}
\usepackage[capitalise]{cleveref}
\crefname{section}{Sec.}{Secs.}  %
\crefname{subsection}{Sec.}{Secs.}  %
\crefname{subsubsection}{Sec.}{Secs.}  %
\crefname{appendix}{App.}{Apps.}  %

\usepackage{graphicx}
\graphicspath{{figures/}}

\usepackage[breakable]{tcolorbox}

\newcounter{takeaway}
\usepackage{tikz}
\usetikzlibrary{calc}
\newlength{\tikzwidth}
\newlength{\boxpad}
\newlength{\boxgap}
\newlength{\boxtextheight}
\title{\THFM: A Unified Video Foundation Model for 4D Human Perception and Beyond}
\usepackage[numbers, sort&compress]{natbib}
\correspondingauthor{my-email@google.com}

\reportnumber{} %

\author[1]{Letian Wang}
\author[1]{Andrei Zanfir}
\author[1]{Eduard Gabriel Bazavan}
\author[1]{Misha Andriluka}
\author[1]{Cristian Sminchisescu$^{*}$}
\affil[]{Google DeepMind}

\begin{abstract}

We present \THFM, a unified video foundation model for human-centric perception that jointly addresses dense tasks (depth, normals, segmentation, dense pose) and sparse tasks (2d/3d keypoint estimation) within a single architecture.
\THFM is derived from a pretrained text-to-video diffusion model, repurposed as a single-forward-pass perception model and augmented with learnable tokens for sparse predictions.
Modulated by the text prompt, our single unified model is capable of performing various perception tasks. 
Crucially, our model is on-par or surpassing state-of-the-art specialized models on a variety of benchmarks despite being trained exclusively on synthetic data (i.e.~without training on real-world or benchmark specific data).
We further highlight intriguing emergent properties of our model, which we attribute to the underlying diffusion-based video representation.
For example, our model trained on videos with a single human in the scene generalizes to multiple humans and other object classes
such as anthropomorphic characters and animals (see Fig.~\ref{fig:oodvis}) -- a capability that hasn't been demonstrated in the past.

\end{abstract}

\begin{document}

\begingroup
\renewcommand{\thefootnote}{}
\footnotetext{\hspace{-1.4mm}$^*$Work done while at Google DeepMind.}
\endgroup

\maketitle

\section*{Introduction}

\begin{figure*}[tb]
   \centering
   \includegraphics[width=1.0\textwidth]{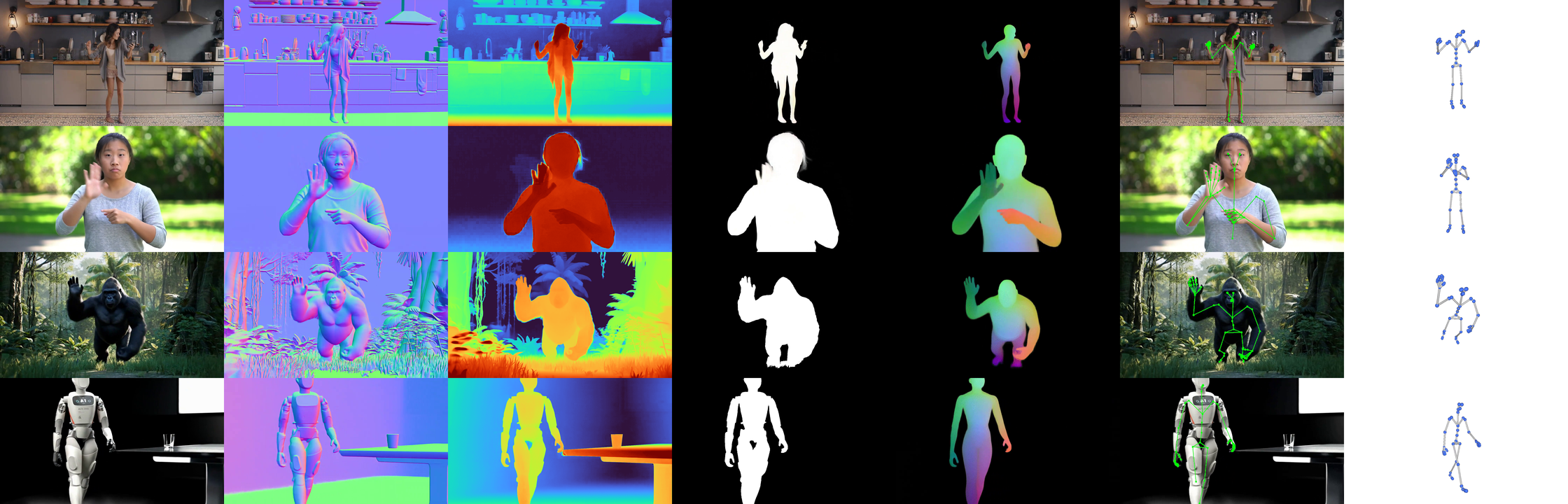}
   \captionof{figure}{\textbf{\THFM is a single, unified video perception model with SOTA performance on a multitude of output modalities}. From left to right we show: a frame from the
   input video, estimated surface normals, depth, segmentation, dense human semantics \cite{wang2019normalized,Guler2018DensePose}, 2d and 3d keypoints. Our approach has been trained on videos of people generated synthetically, yet it
   generalizes to real videos both for people as well as other categories such as
   animals and anthropomorphic characters. 
  \label{fig:teaser}} 
\end{figure*}

In this paper we introduce \THFM\footnote{Temporal Human Foundation Model}, a generalist model for human perception in video
capable of performing on-par or better than state-of-the-art specialized models. Various human perception
tasks have been considered in the literature, ranging from fairly simple pixel-level labeling tasks
such as matting \cite{lin2021real,ke2022modnet,li2021privacy}, to more complex tasks such as dense human semantics \cite{Guler2018DensePose}, recovery of geometric
information in the form of normal and depth maps \cite{kim2025geomantemporallyconsistenthuman}, and reconstruction of articulated human movement
in 2d \cite{sapienseccv2024} and 3d \cite{genmo2025}. Most of these tasks are closely related, yet have been primarily
addressed via specialized single-task models in the past.
As we demonstrate in this paper, recent advances in diffusion-based video modelling enable us to construct a generalist video understanding model following conceptually straightforward techniques.
Our results present strong evidence that modern video diffusion
models (VDM) provide excellent video representation for diverse computer vision tasks. This is exemplified
by several types of generalization results in this paper: (1) our models trained exclusively on
videos of humans perform surprisingly well on categories with substantially different appearance
such as anthropomorphic characters and animals, see Fig.~\ref{fig:teaser}~and~\ref{fig:oodvis}, (2) the fine-grained details produced by our
model at test time are well beyond the level of detail in our synthetic training data, see details
of clothing and hair in Fig.~\ref{fig:teaser}. %

As a starting point for our approach we use a pre-trained text-to-video diffusion model. 
These models are able to generate videos with remarkable
variety of objects and events while maintaining consistency of generated content, all in response to language instructions.
Intuitively, such models must have learned a representation of video that
implicitly captures geometry, appearance, and spatial-temporal evolution of the real world. We aim to make such information explicit via supervised fine-tuning on an appropriate dataset with input/output pairs encoding diverse set of tasks shown on Fig.~\ref{fig:teaser}, while modulating the task via a fixed set of text prompts (see Fig.~\ref{fig:overview}).

\paragraph{Our contributions:} We propose \THFM, a human-centric, unified video foundation model architecture, and mark the following contributions: (1) a unified model capable of outputting multiple dense and sparse modalities, modulated solely by the \textbf{text} condition, exploiting the native architecture of text-to-video diffusion models; (2) we show that a video diffusion backbone can output accurate, temporally smooth and frame aligned geometric estimates in a single forward pass of the model, solely by finetuning on synthetic data; this includes structured arrays of e.g. 3d keypoints %
(3) Crucially, we demonstrate that our unified model achieves state-of-the-art performance across multiple tasks, on-par or surpassing specialized models that are designed for only a single output modality.
(4) we show that video diffusion models, finetuned in this way, exhibit emergent behaviors that go beyond their training set: the predicted modalities remain consistent and plausible for OOD categories (e.g. anthropomorphic characters, animals); 

\section*{Related Work}

\begin{table}[htb]
    \centering
    \resizebox{0.8\linewidth}{!}{%
    \begin{tabular}{lcccccccccc}
        \toprule
        \midrule
         \multirow{2}{*}{Method} &
         \multirow{2}{*}{\shortstack{Number \\ of tasks}} &
         \multirow{2}{*}{\shortstack{One model for \\all tasks}} &
         \multirow{2}{*}{\shortstack{Sparse\\tasks}} &
         \multirow{2}{*}{\shortstack{Task by \\ Text Prompt}} & 
         \multirow{2}{*}{\shortstack{Temporal\\smooth.}} &
         \multirow{2}{*}{\shortstack{1-Step\\Inference}}\\
            
         & & & \\

        \midrule
        \multicolumn{7}{c}{Diffusion-based models (single-frame)} \\
        \midrule
        
        Marigold~\cite{Ke_2024_CVPR} & 1 & \mineno & \mineno & \mineno & \mineno & \mineno \\
        GeoWizard~\cite{fu2024geowizard} & 1 & \mineno & \mineno & \mineno & \mineno & \mineno\\
        Marigold~\cite{ke2025marigold} & 3 & \mineno & \mineno & \mineno & \mineno & \mineno \\
        \citet{martingarcia2024diffusione2eft}  & 1 & \mineno & \mineno & \mineno & \mineno & \mineyes \\
        GenPercept~\cite{xu2024diffusion} & 4 & \mineno & \mineno & \mineno & \mineno & \mineyes \\
        Diception~\cite{zhao2025diception} & 6 & \mineyes & \mineno & \mineyes & \mineno & \mineno \\

        \midrule
        \multicolumn{7}{c}{Diffusion-based models (video)} \\
        
        \midrule

        DepthCrafter~\cite{hu2024depthcraftergeneratingconsistentlong} & 1 & \mineno & \mineno & \mineno & \mineyes & \mineno \\
        NormalCrafter~\cite{bin2025normalcrafterlearningtemporallyconsistent} & 1 & \mineno & \mineno & \mineno & \mineyes & \mineno \\
        DepthAnyVideo~\cite{yang2025depthvideoscalablesynthetic} & 1 & \mineno & \mineno & \mineno & \mineyes & \mineno \\
        Geo4D~\cite{jiang2025geo4d} & 1 & \mineno & \mineno & \mineno & \mineyes & \mineno \\
        DiffusionRenderer~\cite{liang2025diffusionrendererneuralinverseforward} & 3 & \mineyes & \mineno & \mineno & \mineyes & \mineno \\
        \midrule
        \multicolumn{7}{c}{Perception foundation models} \\
        \midrule

        SegmentAnything~\cite{kirillov2023segment} & 1 & \mineno & \mineno & \mineyes & \mineno & \mineyes \\
        DepthAnythingV2~\cite{yang2024depth} & 1 & \mineno & \mineno & \mineno & \mineno & \mineyes \\

        Sapiens~\cite{sapienseccv2024} & 4 & \mineno & \mineyes & \mineno & \mineno & \mineyes \\
        David~\cite{saleh2025david} & 3 & \mineyes & \mineno & \mineno & \mineno & \mineyes \\
        \midrule
        \THFM (ours) & 6 & \mineyes & \mineyes & \mineyes & \mineyes & \mineyes  \\
        \bottomrule
    \end{tabular}
    }
    \caption{Systematic comparison of \THFM to related work across key design dimensions. We compare methods along the number of supported tasks, whether a single unified model handles all tasks, support for sparse prediction tasks (e.g.\ articulated 2d/3d pose estimation, which require continuous predictions outside the pixel space), whether task selection is driven by text prompts, temporal smoothness of predictions across video frames, and single-step (non-iterative) inference capability. To the best of our knowledge, \THFM is the first model that unifies multiple sparse and dense video perception tasks within a single architecture, while simultaneously leveraging text-based task conditioning, producing temporally smooth outputs, and operating in a single forward pass.}
    \label{tab:classification_summary}
\end{table}

\textbf{\textit{Visual Representation Learning.}}
Learning rich and robust visual representations has long been a fundamental problem for enabling a wide range of vision tasks. In the literature, unsupervised and self-supervised learning have become core paradigms for representation learning, as they scale efficiently with large amounts of unlabeled data. Among representative approaches, masked autoencoders~\cite{he2022masked,pathak2016context} (MAE), inspired by masked language modeling~\cite{devlin2019bert,brown2020language}, learn to reconstruct missing image regions and have established a foundational framework for visual pretraining, where the learned representations can be effectively finetuned to various downstream vision tasks. Another prominent paradigm, contrastive learning, aims to align semantically similar samples while separating dissimilar ones. By performing contrastive learning across vision and language modalities, such as CLIP~\cite{radford2021learning} and SigLip~\cite{zhai2023sigmoid}, it learns powerful multimodal representations that unlock a wide range of tasks such as open-vocabulary detection and segmentation. 
We note that representation learning for video remains a crucial problem, presenting numerous challenges such as high computational cost of training and the complexity of modeling temporal dynamics.

\noindent\textbf{\textit{Perception Foundation Models.}}
Perception foundation models have recently attracted significant attention, with models such as Segment Anything Model~\cite{kirillov2023segment} and Depth Anything~\cite{yang2024depth} leveraging massive amounts of data to achieve strong generalization across a wide range of visual scenarios. 
For solely \textbf{human-centric perception}, \textit{Sapiens} \cite{sapienseccv2024} is one of the first foundation models pre-trained on a vast, curated dataset of human images, which significantly boosts performance when fine-tuned for diverse human-centric tasks like pose estimation and segmentation. The \textit{DAViD} approach of \cite{saleh2025david} demonstrates that training exclusively on a smaller, high-fidelity synthetic dataset can achieve state-of-the-art accuracy in human-centric tasks like depth and normal estimation. 
Meanwhile, there have been efforts toward a single unified model capable of handling multiple tasks~\cite{saleh2025david,mizrahi20234m,bachmann20244m}. However, to the best of our knowledge, these existing works operate primarily in the image space, lacking an understanding of the temporal dimension, and a truly unified perception model in the video domain has yet to be realized.

\noindent\textbf{\textit{Adapting Diffusion Models for Perception.}}
Recently, an emerging direction is to leverage the rich features learned in pre-trained diffusion models for perception tasks, a path which our work follows. 
Early work \cite{fu2024geowizard,Ke_2024_CVPR} in this area focused mainly on single-image prediction.
\textit{Marigold} \cite{Ke_2024_CVPR} demonstrated how a pre-trained image diffusion
model like stable diffusion \cite{rombach2022high} can be repurposed through fine-tuning on
synthetic data to act as a %
monocular depth estimator.
\cite{martingarcia2024diffusione2eft} improves the efficiency of prior diffusion-based depth
estimators \cite{Ke_2024_CVPR,fu2024geowizard} by aligning the time step with the noise level,
showing that a simple end-to-end fine-tuning approach can create a single-step, deterministic model
that is both fast and highly accurate. \cite{xu2024diffusion} conducts a systematic study and
identifies key components such as feature injection, decoding mechanisms, and training objectives,
that are critical for successfully adapting pre-trained diffusion models to general dense perception
tasks.

An inherent limitation of these single-image models, however, is their inability to produce temporally consistent predictions when applied to video sequences. 
To address this, \textit{Buffer Anytime}~\cite{kuang2024bufferanytimezeroshotvideo} proposes to adapt the existing image diffusion model to the video domain by incorporating a temporal layer for depth and normal estimation problems. 
Going further, some recent methods directly employ native video diffusion models to address the temporal consistency issues. Under this trend, various tasks and techniques have been explored. \textit{DepthCrafter}~\cite{hu2024depthcraftergeneratingconsistentlong} and \textit{NormalCrafter}~\cite{bin2025normalcrafterlearningtemporallyconsistent} incorporate alignment objectives with large-scale Vision Foundation Models, such as CLIP~\cite{radford2021learning} or DINO~\cite{oquab2023dinov2}, to further facilitate training. 
\cite{yang2025depthvideoscalablesynthetic} explores training with scalable synthetic data to overcome the scarcity of video ground truth data. 
Geo4D~\cite{jiang2025geo4d} adapts video diffusion models for 4D scene reconstruction.
\textit{DiffusionRenderer} \cite{liang2025diffusionrendererneuralinverseforward} proposes a holistic framework that uses video diffusion priors to tackle the dual problems of inverse rendering (estimating G-buffers from video) and forward rendering (generating photorealistic images from G-buffers).

However, these advanced methods primarily concentrate on a single, often dense, vision task. Our key observation is that the text prompt provides a powerful interface to steer the model for diverse tasks, owing to the inherent instruction-following capabilities of text-to-video models. Leveraging this interface, we present the first unified model for video perception tasks, including both dense vision tasks (depth, normal, segmentation, dense pose estimation) and sparse vision tasks (2d/3d keypoints prediction).

\section{\THFM}
\begin{figure*}[tb]
  \centering
   \includegraphics[trim=0cm 4.5cm 0cm 0cm, clip, width=1\textwidth]{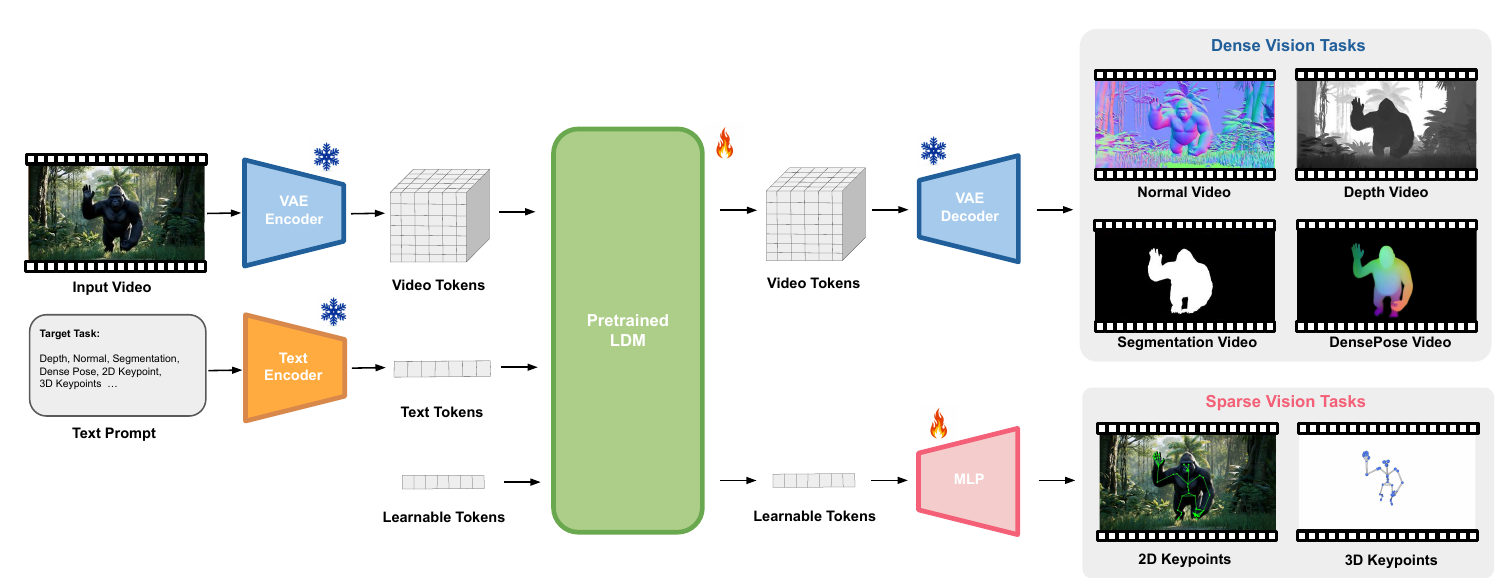}
  \caption{ 
  Method overview of \THFM, a simple yet powerful architecture adapted from text-to-video diffusion models. 
  Given an input video and a text prompting specifying the desired output, our unified model, trained only on synthetic data, is capable of performing a wide range of dense and sparse perception tasks, with a single forward-pass of the model. 
   The dense vision tasks are unified in the RGB ambient space where supervision can be applied in both latent space and RGB ambient space, and the sparse vision tasks are realized by adding learnable tokens as additional inputs to the diffusion transformer (DiT). 
    }

\vspace{-1em}
  \label{fig:overview}
\end{figure*}

{\THFM} is a unified model capable of various human-centric video understanding tasks. As illustrated in Figure~\ref{fig:overview}, we repurpose text-to-video generation models to enable both dense vision tasks (full-scene/human normal estimation, full-scene/human depth estimation, foreground segmentation, dense human semantic estimation) and sparse vision tasks (2d and 3d keypoint prediction) within one single model. 

Given the input RGB video and the text prompt specifying the target modality, our model is able to produce target video in a single forward pass (Sec~\ref{sec: architecture}).
A scalable data synthesis strategy is proposed to enable efficient low-cost collection of diverse high-quality data, which is described in Sec~\ref{sec: data synthesis}.
We then discuss our training recipe in Sec~\ref{sec: training recipe}.

\subsection{Architecture}
\label{sec: architecture}
\textbf{\textit{{Preliminary}.}}
Our model is based on text-to-video diffusion model, with its core components being a VAE encoder-decoder pair, a text encoder, and a transformer-based latent diffusion model (LDM). 
The original LDM model takes a latent token initialized with Gaussian noise and text prompt as inputs and conducts a number of denoising steps, gradually transforming it into latent tokens, which can be decoded to the generated video. The index of the denoising step is provided to LDM as additional input to modulate the predictions with respect to amount of input noise. Let us denote the video token inputs to LDM at step $t$ as $x_t$, with $x_0$ being the noise-free latents of the original video. Depending on the particular formulation, LDM module is trained to predict the noise component $\epsilon$ as in \cite{song2020denoising,ho2020denoisingdiffusionprobabilisticmodels}, or the difference $v = \epsilon - x_0$ between the noise and original input as in the "Rectified flow" variant \cite{liu2022flow, lipman2022flow, esser2024scaling}. 

\noindent\textbf{\textit{From Diffusion to Perception.}} 
Following the best practices in the literature \cite{martingarcia2024diffusione2eft,xu2024diffusion}, we adapt the multi-step diffusion model to a single-step prediction model. To that end we directly use the latent tokens encoded from original video as the inputs to LDM and conduct only a single forward-pass of the model. The input time-step is set to the time step used at the end of multi-step diffusion process, to indicate that the input is noise free. We rely on LDM trained with "Rectified flow" objective, and negate the output of the LDM prior to passing it to the loss functions or decoder. The negated output $-v = x_0 - \epsilon$ is then close to the latent encoding of the RGB video at the start of training, which presumably helps the model to converge faster.

\noindent\textbf{\textit{Unify Dense Tasks in RGB Ambient Space.}} In our unified model, the to-perform task is modulated via the text prompt, where we define a special text conditioning for each target modality. 
The generated latents are fed into the decoder to produce the video of the target modality. 
Note that unlike prior works that adopt separate decoders for different tasks, we adopt a unified approach, where all dense prediction tasks are performed with the same decoder.
To achieve this, the outputs for all dense perception tasks are represented in the RGB space, where the 3 RGB channels are repeated for depth and segmentation tasks that desire a single output dimension, and are responsible for 3 different dimensions for normal and densepose estimation tasks. 

\noindent\textbf{\textit{Enable Sparse Tasks with Learnable Tokens.}} To enable sparse tasks such as 2d and 3d keypoints estimation that are not friendly to RGB representation, we append additional learnable tokens to the video latents, which are together passed as the model inputs, and then a MLP is applied on the additional tokens to produce final keypoint predictions. Specifically, for the video with $T$ frames, we design $T$ learnable tokens, and each token is decoded to predict $K$ keypoints of the human in one frame. 
To indicate the spatial-temporal positions of the additional tokens in a way compliant to the base DiT model, we adopt the 3d RoPE that is natively used in the DiT.
At a high level, 3d RoPE extends vanilla RoPE to indicate positions in time, height and width. For one token $\{\textbf{x}_n\}$, 3d RoPE computes an angle $\theta_n$ based on the token's temporal position $t_n$ and spatial position $h_n, w_n$, and rotates $\textbf{x}_n$ with $\theta_n$ to obtain $\Tilde{\textbf{x}}_n$~\footnote{In fact, RoPE uses a list of angles $\mathbf{\theta} \in \mathcal{R}^{h/2}$ to rotate each element of a vector $\textbf{x}\in \mathcal{R}^h$ separately. We denote it as a single angle for simplicity in the paper.}:
\begin{align}
    \theta_n &= f(t_n, h_n, w_n) \\
    \Tilde{\textbf{x}}_n &= RoPE(\textbf{x}_n, t_n, h_n, w_n) = \textbf{x}_n e^{i\theta_n}
\end{align}
For the original video latents with size $T'\times H\times W$, each of them has a known spatial-temporal position, which pre-defines their 3d RoPE angle. 
For the additional tokens, their spatial position is unknown, thus we apply learnable spatial positions to them. 
Their temporal position is known, while the number of video frames $T$ is longer than the temporal length of video latents $T'$ due to the temporal compression of the encoder.
To avoid going beyond the temporal position seen in training, we apply position interpolation~\cite{chen2306extending} to downscale the temporal position of the additional token. 
We empirically found this additional-query method performs better than the methods that add additional attention layers.

\subsection{Scalable Synthetic Data Generation}
\label{sec: data synthesis}
Training of our model requires high-quality and diverse video data with ground-truth for depth, normal, foreground mask, dense pose, 2d and 3d keypoints. Since most real-world datasets may contain only a subset of the modalities in a limited scale, we resort to synthetic data as a scalable approach to address the data scarcity problem. 

Specifically, we design a synthetic data generation workflow to produce high-quality human-centric video data, covering diverse human entities and motions. We use 800 RenderPeople assets~\cite{Renderpeople.com}, and animate them with 200 motions from the CMU motion capture dataset~\cite{CMUmocap}. 
We use various 3d full scenes or HDRI backdrops similar to \cite{bazavan2022hspacesyntheticparametrichumans} for enhanced background variability, and vary focal lengths, camera positioning and trajectory for enriched camera conditions.
In total, we generate a synthetic dataset of 20,000 videos with various identities, motions and backgrounds. 
To support dense vision tasks, video normals, depth and segmentation masks are generated with separate render passes via Blender \cite{BlenderOnlineCommunity2025}. 
We recover the human joint positions from the rigged \cite{bednarik2024learning} RenderPeople \cite{Renderpeople.com} assets and use them as ground truth for our 2d and 3d keypoint regression task. Each of the generated videos is trimmed to the target number of frames in the video model. In the data preprocessing stage, we encode both the input RGB video and target-modality video into video latents, and the text conditioning prompt into text embeddings.
In each training sample, the model receives two inputs: RGB video latents and the conditioning text embedding. The output modality is variable—it may be the target-modality video, its latents, 2d keypoints, or 3d keypoints—a choice dictated by the specific task and the corresponding training losses, which are addressed in the subsequent section.

\begin{table}[tb!]
    \centering
    \begin{tabular}{ccc}
        \includegraphics[width=0.3\textwidth]{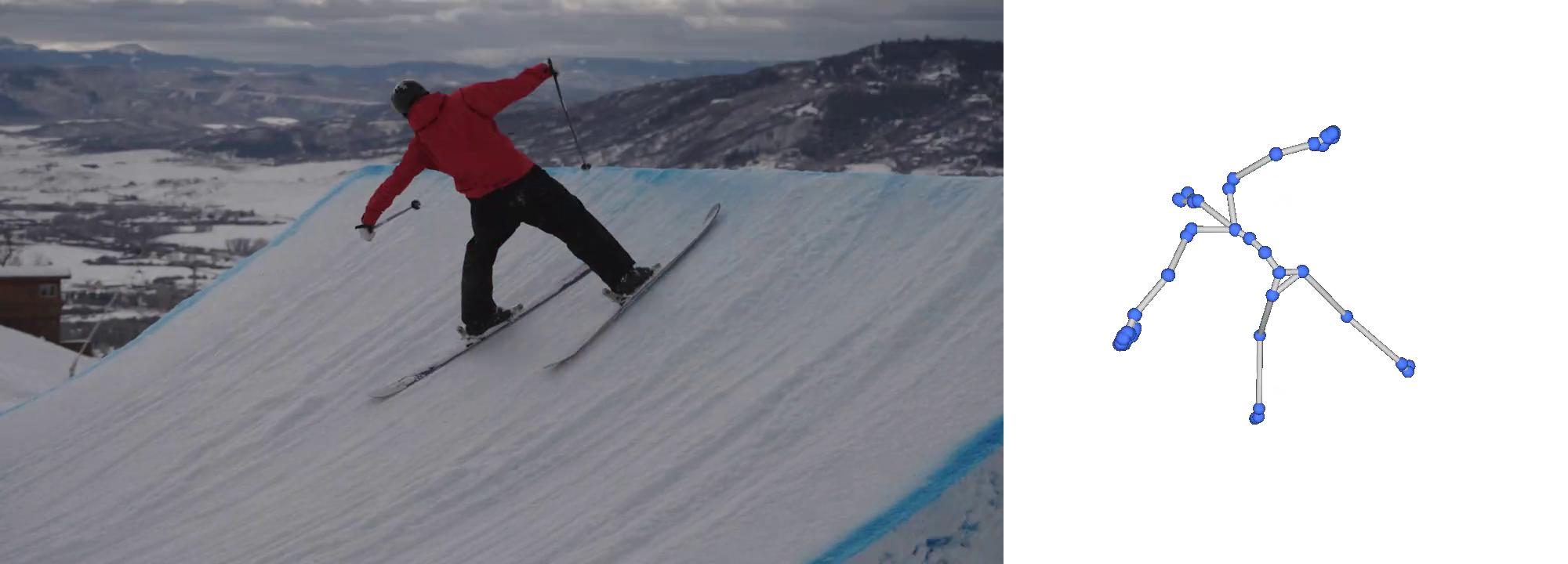} & 
        \includegraphics[width=0.3\textwidth]{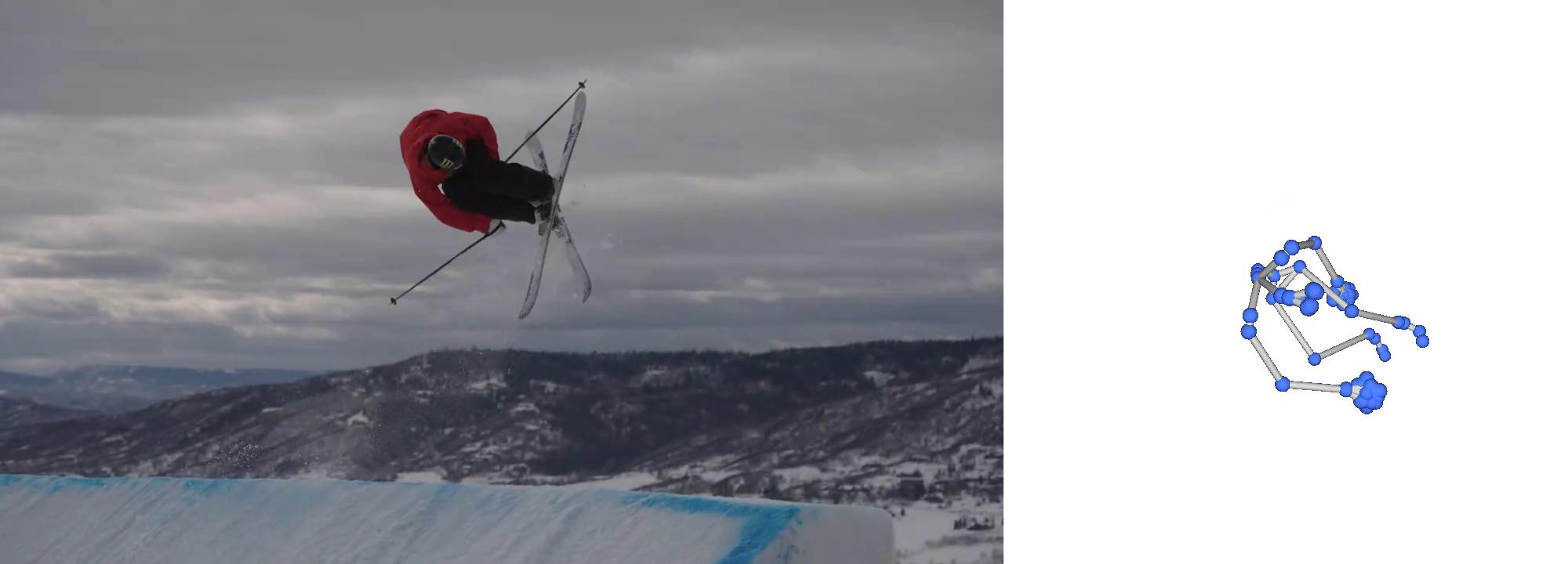} & 
        \includegraphics[width=0.3\textwidth]{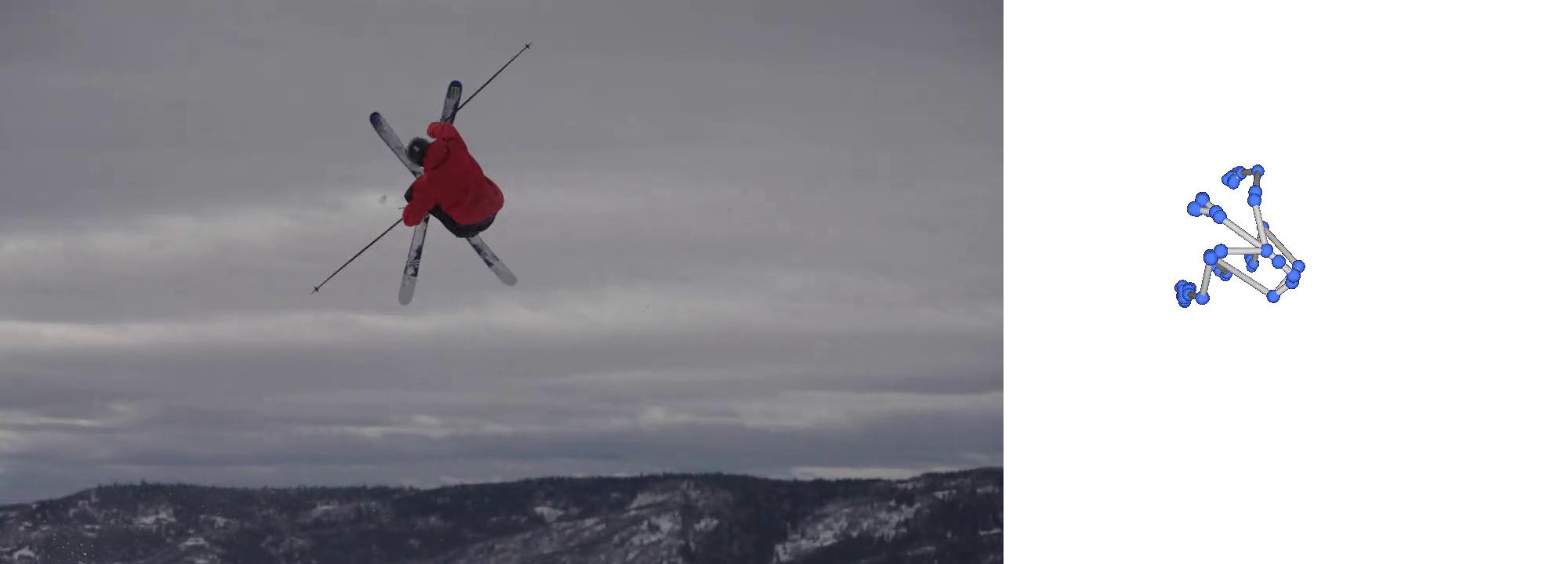} \\
                
        \includegraphics[width=0.3\textwidth]{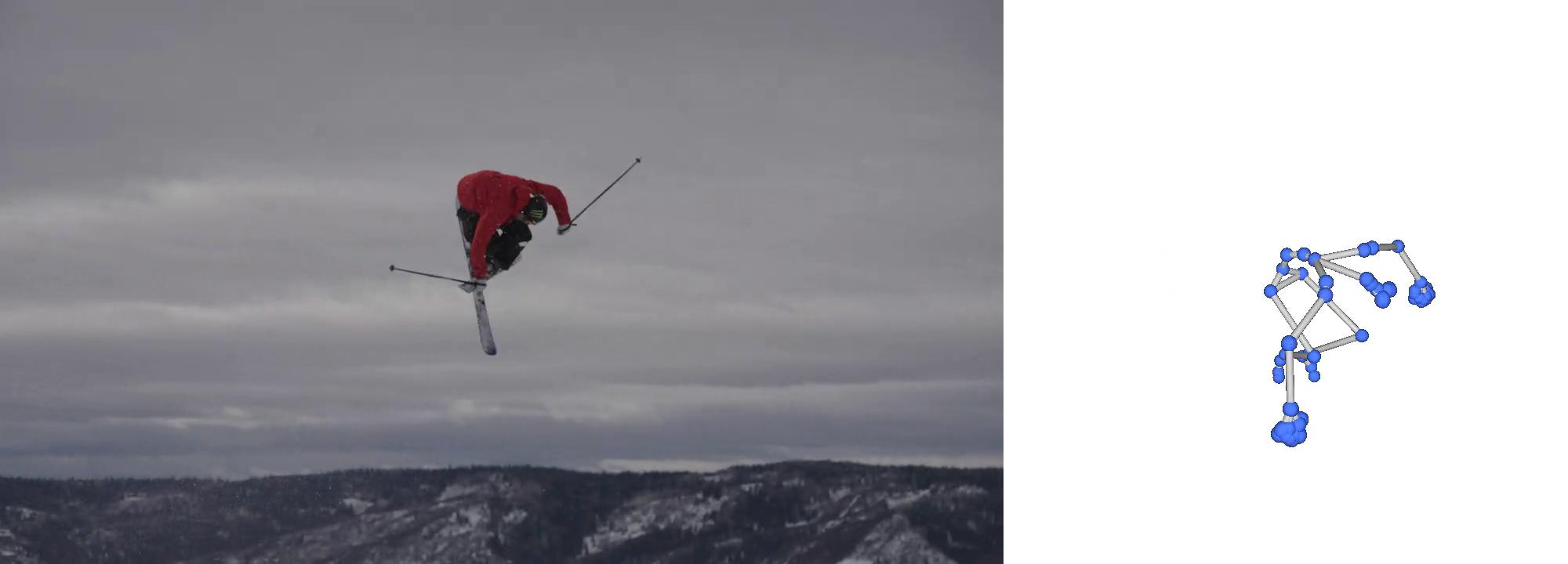} & 
        \includegraphics[width=0.3\textwidth]{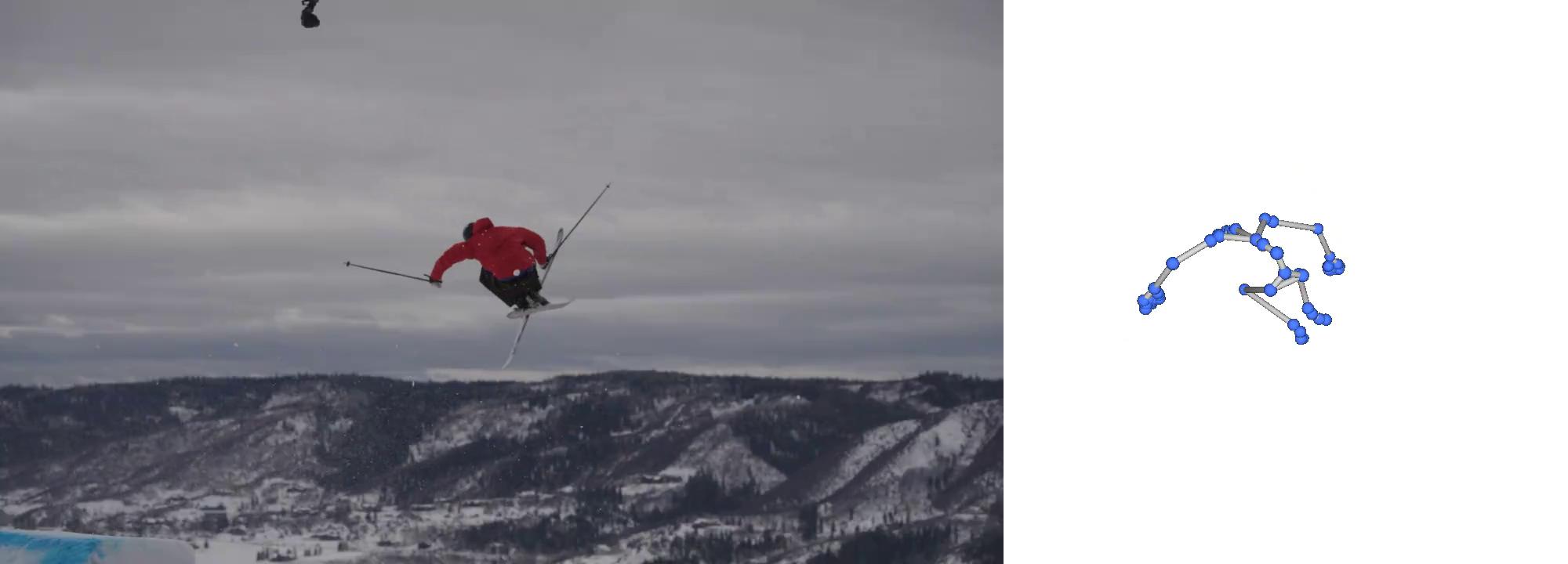} & 
        \includegraphics[width=0.3\textwidth]{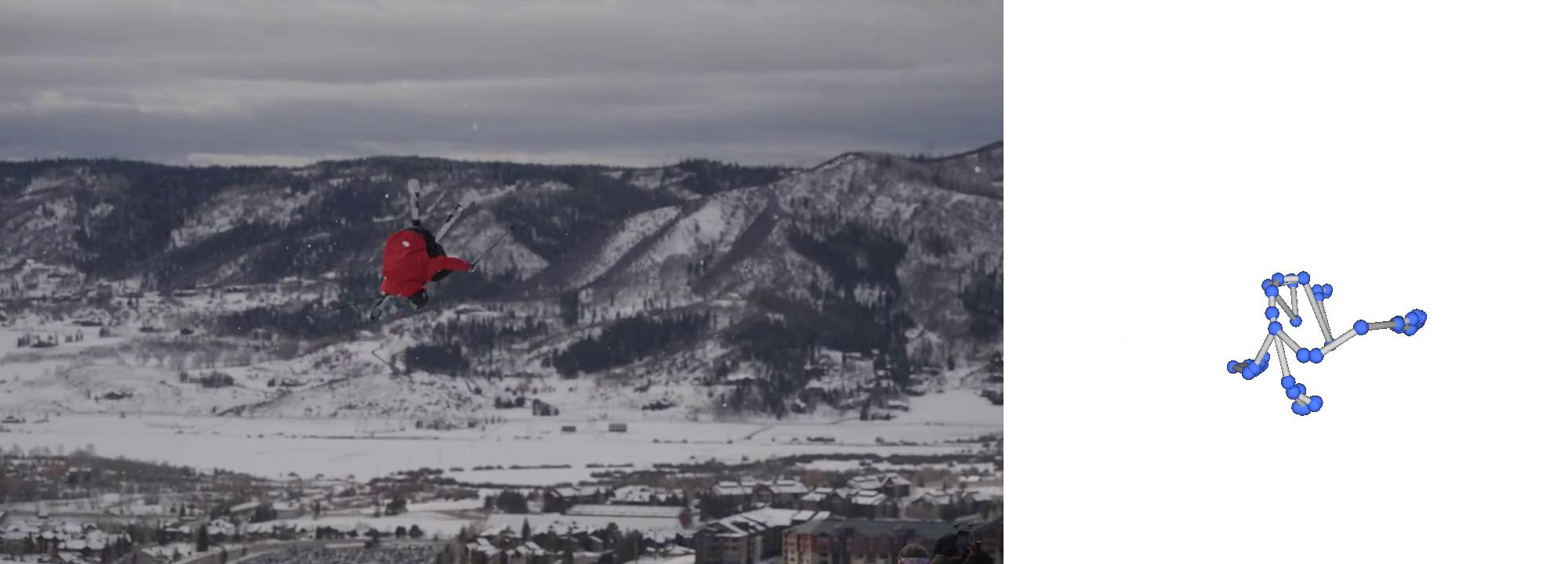} \\  
        
        \includegraphics[width=0.3\textwidth]{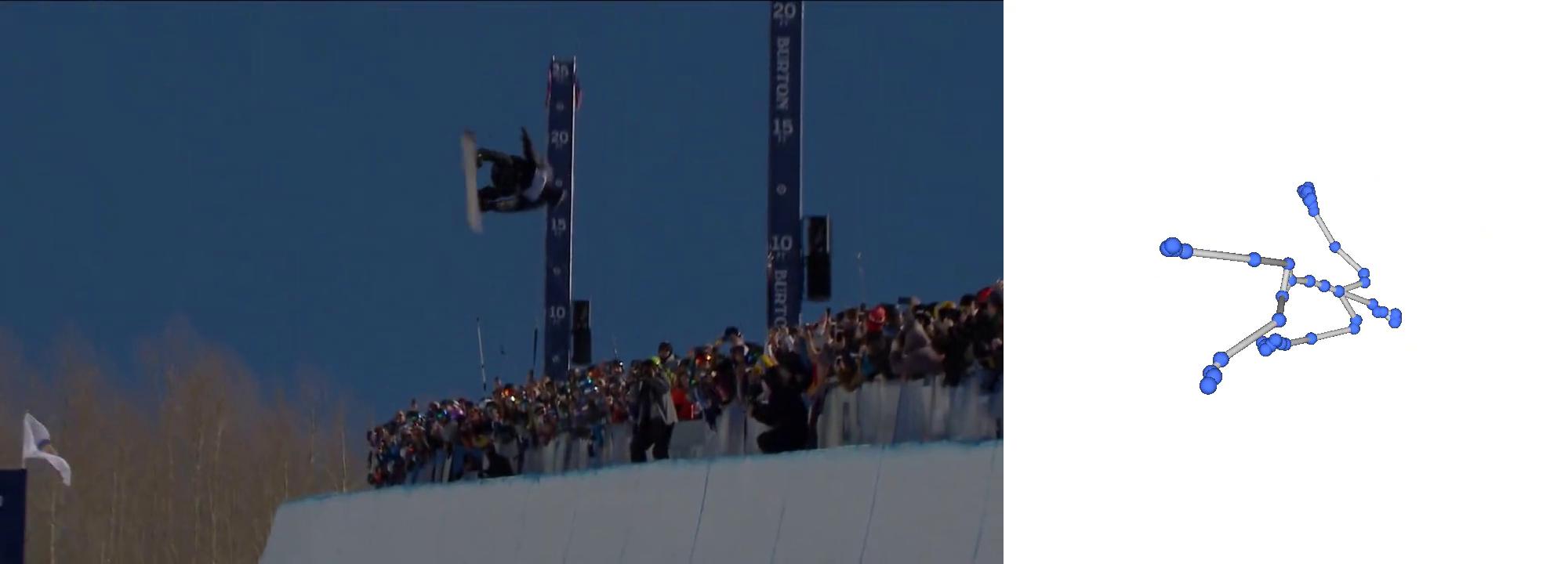} & 
        \includegraphics[width=0.3\textwidth]{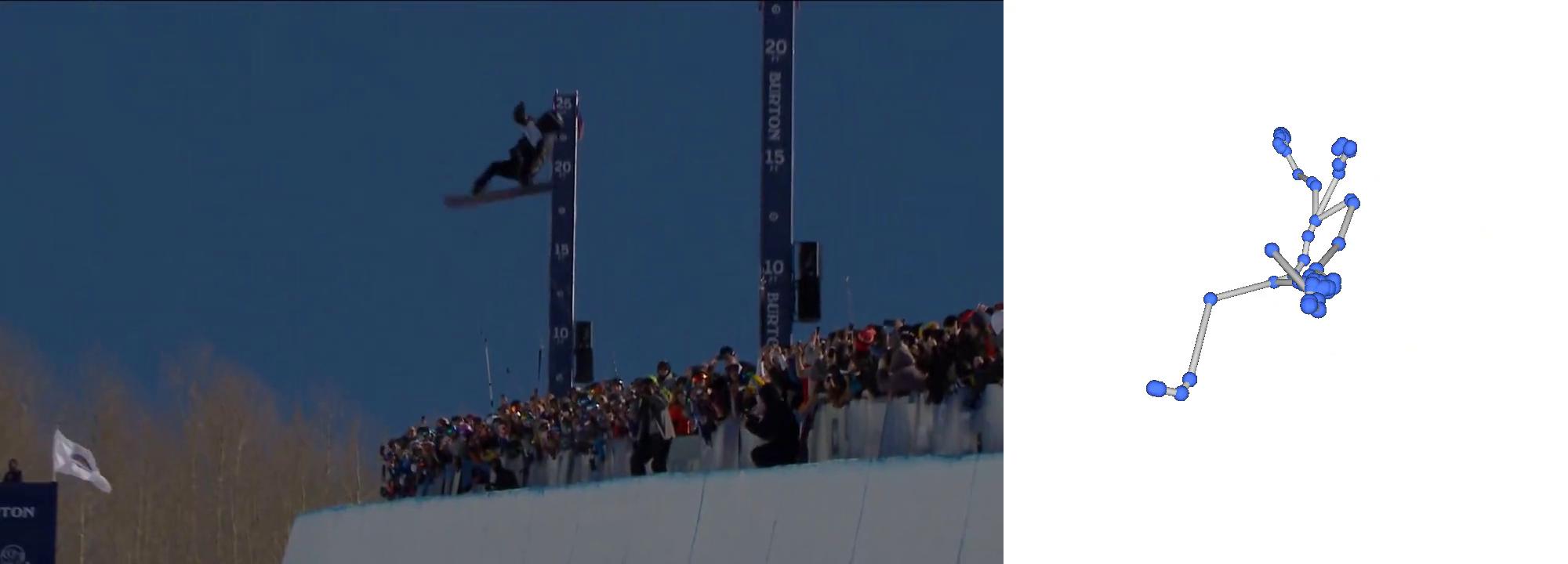} & 
        \includegraphics[width=0.3\textwidth]{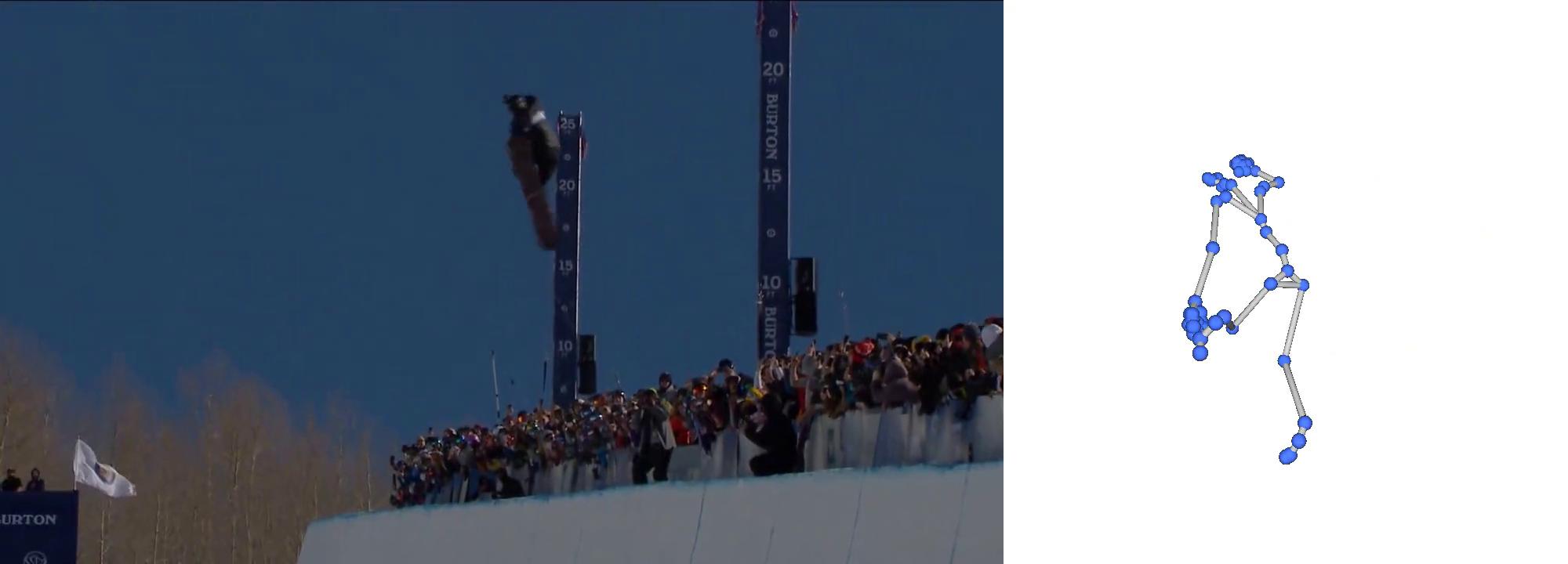} \\

        \includegraphics[width=0.3\textwidth]{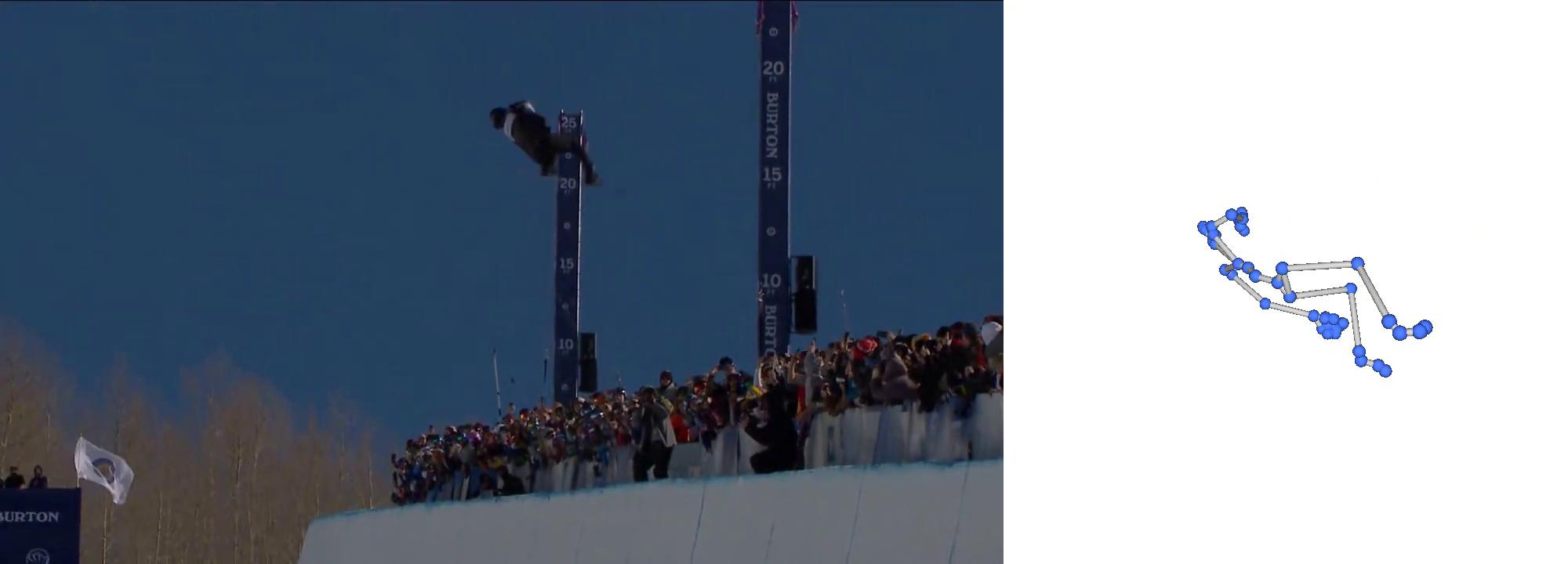} & 
        \includegraphics[width=0.3\textwidth]{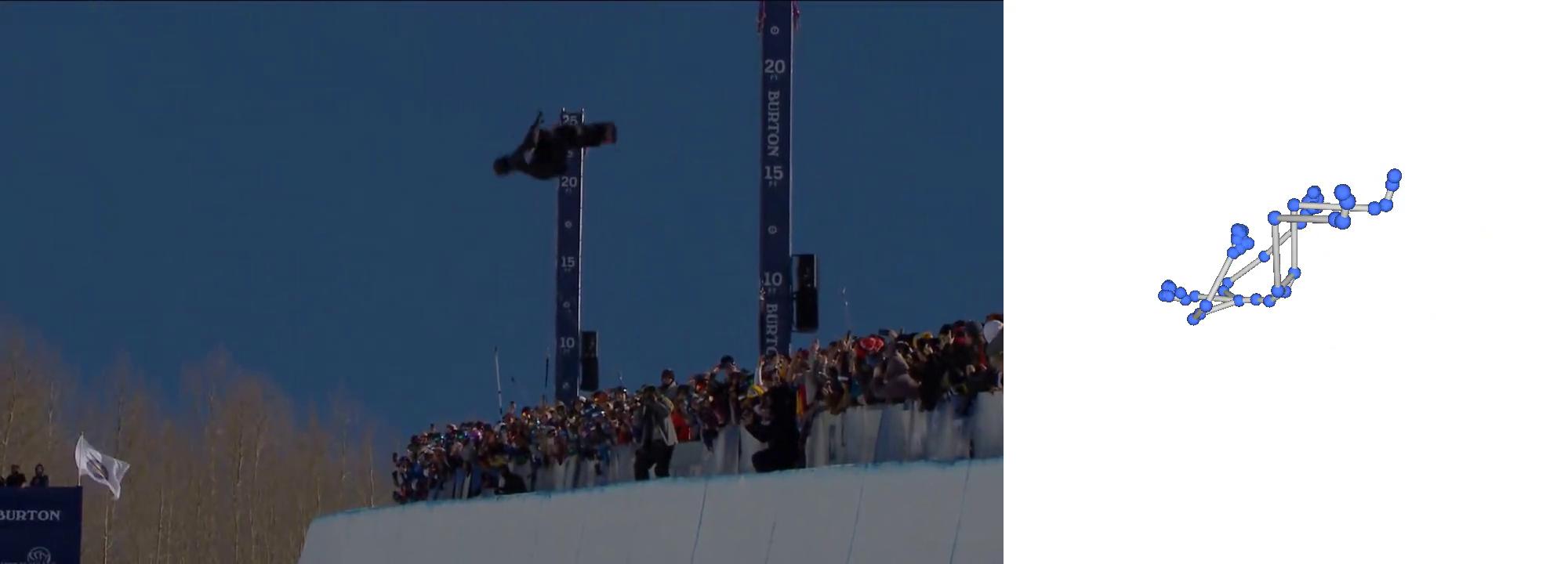} & 
        \includegraphics[width=0.3\textwidth]{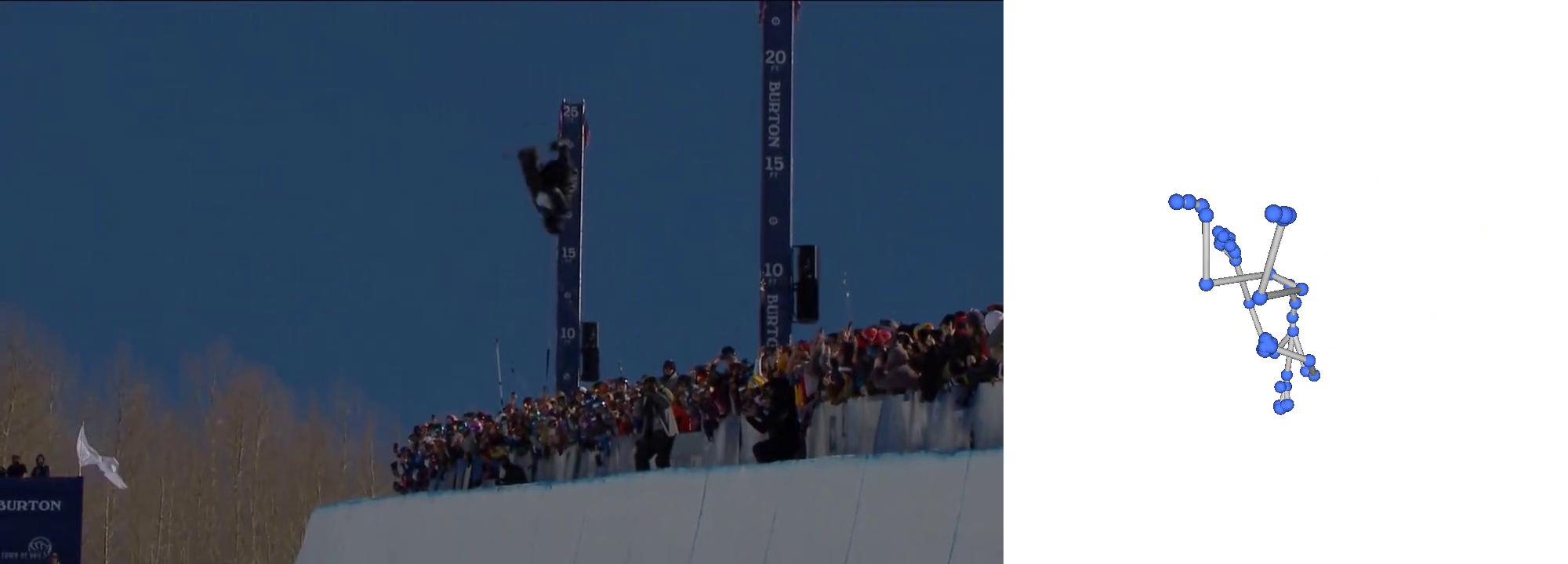} \\
      
    \end{tabular}
    \captionof{figure}{\textbf{3D Pose Estimation}. Example 3D pose estimation results on challenging skiing (top) and snowboarding (bottom) video. We show every 10-th frame for a subsequence of the input video. Compared to the existing state-of-the-art 3D human pose estimation methods \THFM model takes a full video frame as input and does not require pre-processing steps such as person detection or 2d keypoint estimation.}
    \label{fig:skiing}
\end{table}

\subsection{Training Recipe}
\label{sec: training recipe}

With the model architecture established, we now describe the training losses employed for our unified framework.

\noindent\textbf{\textit{Two-Stage Training Strategy.}} For regression tasks, specifically 2d and 3d keypoint prediction, the losses are computed directly on the model's outputs. Conversely, for dense vision tasks, the immediate outputs of the DiT model reside in the latent space and subsequently decoded to the final RGB video space.
While computing losses directly on the latent representations is feasible, we found improved performance by including the decoder in the training loop and employing task-specific losses within the RGB ambient space. This ambient-space strategy offers a significantly richer learning signal by imposing high-resolution, pixel-wise, and task-customized supervision, rather than relying on compressed latent features. This gain in performance, however, comes at the cost of higher memory consumption and a reduced training speed.
Consequently, we adopt a two-stage training strategy: we first train the model in the latent space with mean squared loss to achieve reasonable performance efficiently and quickly, and then continue training in the ambient space to leverage the enriched learning signal in modality-specific losses.
In the following, we will discuss the customized loss for each modality.

\noindent\textbf{\textit{Surface Normal Estimation.}} The 3 channels in the output RGB video are designed to correspond to the \textit{xyz} components of the normal vector at each pixel. Our model is trained to maximize the alignment between predicted vector $\textbf{y}$ and ground-truth vector $\hat{\textbf{y}}$: 
\begin{equation}
    \mathcal{L}_{normal} = ||\textbf{y} - \hat{\textbf{y}}||_2 + (1 - \frac{\textbf{y} \cdot \hat{\textbf{y}}}{|\textbf{y}||\hat{\textbf{y}}|})
\end{equation}

\noindent\textbf{\textit{Depth Estimation.}} 
For depth estimation, following prior arts~\cite{yang2024depth,hu2024depthcraftergeneratingconsistentlong}, our model predicts depth in the disparity space. Specifically, we first transform ground-truth depth $z$ into the disparity space by $d = 1 / z$, and then normalize the ground-truth disparity to the range of [0, 1] by the maximum and minimum disparity values in the video sequence. We then adopt scale-and-shift invariant loss~\cite{ranftl2020towards} to be robust to scales and shift:
\begin{equation}
    \mathcal{L}_{depth} = \frac{1}{THW}\sum_{i=1}^{THW}\rho(d_i, \hat{d_i}), 
\end{equation}
where $d_i$ and $\hat{d_i}$ are the ground truth and prediction, respectively. And $\rho$ is the scale-and-shift invariant root mean squared error loss: $\rho(d_i, \hat{d_i}) = |d_i^* - \hat{d}_i^*|_2$, where $d^*_i$ and $\hat{d}^*_i$ are the scaled and shifted versions of the ground truth $d_i$ and prediction $\hat{d}_i$, respectively:
\begin{equation}
    d_i^* = \frac{d_i - t(d)}{s(d)}, \hspace{2em}
    \hat{d}_i^* = \frac{\hat{d}_i - t(\hat{d})}{s(\hat{d})},
\end{equation}
where $t(d)$, $s(d)$, $t(\hat{d})$, and $s(\hat{d})$ denote the median and mean absolute deviation of the prediction $d$ and ground-truth $\hat{d}$ respectively, which are designed to align the prediction and ground truth to have zero translation and unit scale.
Note that, our disparity prediction shares the same normalization, scale and shift across frames, rather
than a per-frame setup, providing enhanced temporal consistency.

\noindent \textbf{\textit{Other Tasks.}} For the remaining tasks, including dense pose estimation, foreground segmentation, and 2d/3d keypoints, we found a straightforward $L_2$ loss delivered competitive performance.

\section{Experiments}
We comprehensively evaluate our model on various challenging real-world datasets across four tasks: depth estimation, normal estimation, foreground instance segmentation, and 3d keypoint prediction.

\subsection{Implementation Details}
\label{sec: implementation details}

Our approach is in principle agnostic to a particular implementation of the latent diffusion backbone.
Most of our experiments are based on an open-source and open-weights text-to-video diffusion model WAN~\cite{wan2025}. 
Following its original architecture, we train our model on video samples at a resolution of 480x832, comprising 81 frames captured at 24 frames per second (FPS). The model's encoder downsamples the input dimensionality with a temporal factor of 4 and a spatial factor of 8 for both width and height.
The training proceeds in two stages. In stage 1, we train the model in the latent space for 20,000 steps with a batch size of 64 on 256 v6e TPUs . In stage 2, we continue training our model in the ambient RGB space for an additional 10,000 steps with a reduced batch size of 16 on 64 v4 TPUs.
We optimize the model using the Adam optimizer~\cite{kingma2014adam}, with a learning rate of $5 \times 10^{-5}$, implementing a linear warmup over the first $10\%$ of the training steps.
Crucially, for training stability, we apply both gradient clipping, constraining the gradient norm within a certain threshold, and gradient dropping, which discards any batch whose gradient norm exceeds a higher threshold. These techniques proved vital for achieving high-quality performance.

\begin{table}[!t]
\centering
\scriptsize
\resizebox{0.9\linewidth}{!}{%
\begin{tabular}{l| cc c}
\toprule
\multirow{3}{*}{Method} & 
\multicolumn{3}{c}{Hi4D} \\ 
\cmidrule(lr){2-4} 
& \multicolumn{2}{c}{Angular Error (°) $\downarrow$} & \% Within $t^\circ$ $\uparrow$  \\
 \cmidrule(lr){2-3} \cmidrule(lr){4-4}

& Mean & Median & 11.25° / 22.5° / 30° \\
\midrule

PIFuHD \cite{saito2020pifuhd} & 22.39 & 19.26 & 23.0 / 60.1 / 77.0 \\
HDNet \cite{jafarian2021learning}  & 28.60 & 26.85 & 19.1 / 57.9 / 70.1 \\
ICON \cite{xiu2022icon}   & 20.18 & 17.52 & 26.8 / 66.3 / 82.7 \\
ECON \cite{xiu2023econ}  & 18.46 & 16.47 & 29.3 / 68.1 / 84.9 \\
David-Base \cite{saleh2025david} & 15.72 & 12.95 & 43.2 / 78.7 / 89.2\\
David-Large \cite{saleh2025david} & 15.37 & 12.51 & 45.1 / 79.7 / 89.6\\
Sapiens-0.3B \cite{sapienseccv2024} & 15.04 & 12.22 & 47.1 / 81.5 / 90.7\\
Sapiens-2B \cite{sapienseccv2024} &  \underline{12.14} & \underline{9.62} & \underline{60.2 / 89.1} / 94.7\\
\midrule

Decoder \hspace{0em}\textbar\hspace{0.8em} 
Input \hspace{0.8em}\textbar\hspace{0em} 
4 Dense \hspace{0.0em}\textbar\hspace{0em} 
2 Sparse \hspace{0.0em}\textbar\hspace{0em} 
Token & 
\multicolumn{3}{c}{\multirow{2}{*}{{\THFM} Variant}} \\
Training \hspace{0em}\textbar\hspace{0em} 
Timestep \hspace{0em}\textbar\hspace{0.5em} 
Task \hspace{1em}\textbar\hspace{0.7em} 
Task \hspace{1.1em}\textbar\hspace{0.3em} 
Mask \\
\midrule

\hspace{1em} \mineno \hspace{3.2em} End \hspace{2.8em} \mineno \hspace{3.4em} \mineno \hspace{2.8em} \mineno & 13.04 & 10.80 & 52.4 / {88.5} / {94.8} \\

\hspace{1em} \mineno \hspace{3.1em} Start \hspace{2.7em} \mineyes \hspace{3.3em} \mineno \hspace{2.8em} \mineno & 13.62 & 10.67 & 52.9 / {85.8} / {92.9}\\

\hspace{1em} \mineno \hspace{3.2em} End \hspace{2.9em} \mineyes \hspace{3.3em} \mineno \hspace{2.8em} \mineno & 12.26 & 9.69 & 58.4 / \underline{89.1 / 94.8} \\

\hspace{1em} \mineyes \hspace{3.2em} End \hspace{2.9em} \mineyes \hspace{3.3em} \mineno \hspace{2.8em} \mineno & \textbf{11.01} & \textbf{8.96} & \textbf{64.0} / \textbf{92.4} / \textbf{96.5}\\

\hspace{1em} \mineyes \hspace{3.2em} End \hspace{2.9em} \mineyes \hspace{3.3em} \mineyes \hspace{2.8em} \mineno & {13.47} & {10.63} & {53.1} / {86.0} / {93.1}\\

\hspace{1em} \mineyes \hspace{3.2em} End \hspace{2.9em} \mineyes \hspace{3.3em} \mineyes \hspace{2.8em} \mineyes & {13.30} & {11.19} & {50.2} / {88.2} / {94.7}\\

\bottomrule
\end{tabular}%
}
\vspace{-1em}
\caption{Surface normal estimation results on the Hi4D~\cite{yin2023hi4d} dataset. We follow the evaluation protocol of Sapiens~\cite{sapienseccv2024} and DAVID~\cite{saleh2025david}. We report mean and median angular error (lower is better) alongside the percentage of pixels within angular thresholds $t \in \{11.25\degree, 22.5\degree, 30\degree\}$ (higher is better). Baseline results are taken from DAVID~\cite{saleh2025david,sapienseccv2024}. The bottom rows present ablations of \THFM across key design choices: decoder training space (latent vs.\ ambient), input timestep selection, number of jointly trained tasks, and token masking. %
}
\label{tab:normal_eval_all}
\end{table}

\begin{table}[!t]
\resizebox{0.85\linewidth}{!}{%
\begin{tabular}{l|cccccccccc}
\toprule
\multirow{2}{*}{Method} & 
\multicolumn{2}{c}{Goliath-Face} & 
\multicolumn{2}{c}{Goliath-UpperBody} & 
\multicolumn{2}{c}{Goliath-FullBody} & 
\\
\cmidrule(lr){2-3} \cmidrule(lr){4-5} \cmidrule(lr){6-7} \cmidrule(lr){8-9}
& RMSE $\downarrow$ & AbsRel $\downarrow$ & RMSE $\downarrow$ & AbsRel $\downarrow$ & RMSE $\downarrow$ & AbsRel $\downarrow$\\
\midrule

\multicolumn{7}{c}{Results from DAVID} \\
\midrule
MiDaS-DPT\_L~\cite{ranftl2020towards} & 0.224 & 0.016  & 0.553 & 0.015 & 0.973 & 0.027 \\
 DepthAnythingV2-L~\cite{yang2025depth} & 0.229 & 0.017& 0.492 & 0.014 & 1.039 & 0.029 \\
Depth-Pro~\cite{bochkovskii2024depth} & 0.295 & 0.020 & 0.442 & 0.010 & 0.723 & 0.016\\
Sapiens-0.3B~\cite{sapienseccv2024} & 0.179 & \underline{0.012} & \underline{0.368} & 0.010 & 0.690 & 0.019 \\
 Sapiens-2B~\cite{sapienseccv2024} & 0.158 & \textbf{0.009} & \textbf{0.204} & \textbf{0.005} & \textbf{0.266} & \textbf{0.007} &  \\
DAVID-Base~\cite{saleh2025david} & \underline{0.142} & \textbf{0.009} & 0.316 & 0.009 & 0.376 & 0.010 \\
DAVID-Large~\cite{saleh2025david} & \textbf{0.140} & \textbf{0.009} & {0.283} & \underline{0.008} & \underline{0.334} & \underline{0.009} \\
\midrule

\multicolumn{7}{c}{Our Eval} \\
\midrule
Depth-Pro~\cite{bochkovskii2024depth} & 0.034 & 0.023 & 0.060 & 0.015 & 0.104 & 0.028 \\
Sapiens-0.3B~\cite{sapienseccv2024} & {0.023} & {0.013} & \underline{0.038} & \underline{0.009} & 0.096 & 0.026 \\

Sapiens-2B~\cite{sapienseccv2024} & \underline{0.022} & \textbf{0.011} & \textbf{0.028} & \textbf{0.006} & {0.049} & {0.011} &  \\

DAVID-Base~\cite{saleh2025david}  & \underline{0.022} & \underline{0.012} & {0.043} & {0.011} & {0.076} & {0.020} \\

DAVID-Large~\cite{saleh2025david}  & \underline{0.022} & \underline{0.012} & \underline{0.038} & \underline{0.009} & {0.058} & {0.014} \\
\midrule

\midrule

Decoder \hspace{0em}\textbar\hspace{0em} 
4 Dense \hspace{0.0em}\textbar\hspace{0em} 
2 Sparse \hspace{0.0em}\textbar\hspace{0em} 
Token & 
\multicolumn{6}{c}{\multirow{2}{*}{{\THFM} Variant}} \\
Training \hspace{0em}\textbar\hspace{0.7em} 
Task \hspace{0.8em}\textbar\hspace{0.7em} 
Task \hspace{1.1em}\textbar\hspace{0.2em} 
Mask \\
\midrule

\hspace{1.2em} \mineyes \hspace{3.5em} \mineno \hspace{3.5em} \mineno \hspace{3.3em} \mineno & \textbf{0.021} & \textbf{0.011} & \textbf{0.028} & \textbf{0.006} & \textbf{0.037} & \textbf{0.007} \\

\hspace{1.2em} \mineyes \hspace{3.5em} \mineyes \hspace{3.5em} \mineno \hspace{3.3em} \mineno & {0.026} & {0.016} & \underline{0.038} & \underline{0.009} & {0.043} & \underline{0.009} \\

\hspace{1.2em} \mineyes \hspace{3.5em} \mineyes \hspace{3.5em} \mineyes \hspace{3.3em} \mineyes & {0.033} & {0.023} & \underline{0.038} & \underline{0.009} & \underline{0.042} & \underline{0.009} \\

\bottomrule
\end{tabular}
}
\centering
\footnotesize
\vspace{-1em}
\caption{Depth estimation results on the Goliath~\cite{martinez2024codec} dataset, evaluated across three body-crop settings (Face, UpperBody, FullBody) using RMSE and AbsRel metrics (lower is better). We report two groups of results: ``Results from DAVID'' reproduces numbers as reported in DAVID~\cite{saleh2025david}; ``Our Eval'' provides our re-evaluation of publicly available models using the same data splits (12 cameras, 16 frames, 4 subjects) to ensure a consistent comparison protocol. The bottom rows show ablations of \THFM variants. %
}
\label{tab:depth_eval}
\end{table}

\begin{table}[t]
\begin{scriptsize}

\begin{center}
\begin{tabular}{l|cccc}
\toprule
\multirow{2}{*}{Method} & \multicolumn{4}{c}{VideoMatte} \\
\cmidrule(lr){2-5} 
& MAD & MSE & Grad & dtSSD \\
 \cmidrule(lr){1-1} \cmidrule(lr){2-5}
MODNet + FGF~\cite{ke2022modnet} &  11.04 & 5.42 & 15.80 & 3.10 \\
RVM~\cite{lin2022robust}  & \underline{5.64} & \underline{1.07} & \underline{9.80} & \textbf{1.84} \\

\midrule

\midrule
Decoder \hspace{0em}\textbar\hspace{0em} 
4 Dense \hspace{0.0em}\textbar\hspace{0em} 
2 Sparse \hspace{0.0em}\textbar\hspace{0em} 
Token & 
\multicolumn{4}{c}{\multirow{2}{*}{{\THFM} Variant}} \\
Training \hspace{0em}\textbar\hspace{0.7em} 
Task \hspace{0.8em}\textbar\hspace{0.7em} 
Task \hspace{1.1em}\textbar\hspace{0.2em} 
Mask \\
\midrule

\hspace{1em} \mineno \hspace{3.4em} \mineyes \hspace{3.2em} \mineno \hspace{2.8em} \mineno & {6.46} & {2.14} & {0.945} & {3.31} \\

\hspace{1em} \mineyes \hspace{3.4em} \mineyes \hspace{3.2em} \mineno \hspace{2.8em} \mineno & \textbf{5.24} & \textbf{1.00} & \textbf{0.458} & \underline{2.80} \\

\hspace{1em} \mineyes \hspace{3.4em} \mineyes \hspace{3.2em} \mineyes \hspace{2.8em} \mineno & {6.97} & {2.69} & {1.41} & {3.56} \\

\hspace{1em} \mineyes \hspace{3.4em} \mineyes \hspace{3.2em} \mineyes \hspace{2.8em} \mineyes & {6.01} & {1.75} & {0.938} & {3.33} \\

\bottomrule
\end{tabular}
\end{center}
\vspace{-2.5em}

\begin{center}
\begin{tabular}{l|ccccc}
\toprule
\multirow{2}{*}{Method} & \multicolumn{2}{c}{PhotoMatte85} & \multicolumn{3}{c}{PPM-100}\\
\cmidrule(lr){2-3}  \cmidrule(lr){4-6}
& MSE & Conn & MSE & MAD & Conn \\
 \cmidrule(lr){1-1} \cmidrule(lr){2-3} \cmidrule(lr){4-6}
BSHM~\cite{Liu_2020_CVPR} &  - & - & \underline{0.0063} & \underline{0.0114} & - \\
P3M-Net~\cite{li2021privacy} &  0.0070 & 19.76 & - & - & 139.89 \\
DAVID~\cite{saleh2025david}  & \textbf{0.0009} & {5.6}  & - & - & 74.72 \\
MODNet~\cite{ke2022modnet}  & 0.0030 & 11.18  & \textbf{0.0044} & \textbf{0.0086} & 96.45\\

\midrule

Decoder \hspace{0em}\textbar\hspace{0em} 
4 Dense \hspace{0.0em}\textbar\hspace{0em} 
2 Sparse \hspace{0.0em}\textbar\hspace{0em} 
Token & 
\multicolumn{4}{c}{\multirow{2}{*}{{\THFM} Variant}} \\
Training \hspace{0em}\textbar\hspace{0.7em} 
Task \hspace{0.8em}\textbar\hspace{0.7em} 
Task \hspace{1.1em}\textbar\hspace{0.1em} 
Mask \\
\midrule

\hspace{1em} \mineno \hspace{3.4em} \mineyes \hspace{3.2em} \mineno \hspace{2.8em} \mineno & \underline{0.0014} & \underline{0.80} & {0.0083} & {0.0125} & \textbf{3.07} \\

\hspace{1em} \mineyes \hspace{3.4em} \mineyes \hspace{3.2em} \mineno \hspace{2.8em} \mineno & {0.0016} & {0.87} & \textbf{0.0044} & \textbf{0.0086} & \textbf{2.00} \\

\hspace{1em} \mineyes \hspace{3.4em} \mineyes \hspace{3.2em} \mineyes \hspace{2.8em} \mineno & {0.0019} & {1.00} & {0.0421} & {0.0463} & {12.34} \\

\hspace{1em} \mineyes \hspace{3.4em} \mineyes \hspace{3.2em} \mineyes \hspace{2.8em} \mineyes & \textbf{0.0009} & \textbf{0.61} & {0.0260} & {0.0030} & {8.06} \\

\bottomrule
\end{tabular}
\end{center}
\vspace{-2.5em}

\caption{Evaluation of our approach in soft foreground segmentation on PhotoMatte85~\cite{lin2021real}, PPM100~\cite{ke2022modnet}, and VideoMatte~\cite{lin2021real} dataset (static split). PhotoMatte85 and Conn in PPM-100 taken from DAVID~\cite{saleh2025david} and MODNet~\cite{ke2022modnet}. VideoMatte results taken from RVM~\cite{lin2022robust}.}
\label{tab:evalsoftforeground}

\end{scriptsize}
\end{table}

\subsection{Evaluation Protocol}
\textbf{\textit{Evaluation Datasets.}} We evaluate our approach on a set of challenging, real-world benchmarks that encompass both video and image data. For surface normal estimation, we evaluate on Hi4D datasets~\cite{yin2023hi4d}. Specifically, we adhere to the evaluation protocol established in Sapiens~\cite{sapienseccv2024} and DAVID~\cite{saleh2025david}, selecting the same sequences from pairs 28, 32, and 37. These sequences feature six unique subjects captured by camera 4, resulting in total of 1,195 frames for testing. 
For depth estimation, we follow DAVID~\cite{saleh2025david} to perform evaluation on the Goliath datasets~\cite{martinez2024codec}. To ensure fair comparison, we utilize the identical selection of 12 cameras, 16 frames, and 4 subjects as in DAVID~\cite{saleh2025david}, which are categorized into data splits of face, upper and full body, with approximately 2.2k frames in total.
For soft foreground segmentation, we report results on VideoMatte~\cite{lin2021real}, PhotoMatte 85~\cite{lin2021real}, and PPM-100~\cite{ke2022modnet}. The original VideoMatte dataset does not provide official split, thus we evaluate on the static split composite from \citep{lin2022robust}. For 3D keypoint prediction, we assess our model on EMDB~\cite{kaufmann2023emdb}, RICH~\cite{huang2022cap} and H3.6M\cite{h36m_pami}.
Note that our method is trained solely with synthetic data and has not been exposed to training sets often provided alongside the evaluation benchmarks. %
Also note that all compared methods are specialized in single task, while we show one model capable of multiple tasks.

\noindent\textbf{\textit{Evaluation Metrics.}} 
To assess our model's performance, we report standard metrics appropriate for each task. 
For surface normal estimation, we report the common metrics~\cite{saleh2025david,sapienseccv2024} including mean and median angular error, alongside the percentage of pixels where the error is within the threshold $t \in \{11.25\degree, 22.5\degree, 30\degree\}$. 
For depth estimation, we follow standard practice~\cite{sapienseccv2024,ranftl2020towards,yang2024depth} by adopting the root mean square error (RMSE) and the mean absolute value of the relative depth (AbsRel). 
For soft foreground segmentation, we report common metrics used in prior work~\cite{saleh2025david,yang2025matanyone,lin2022robust,lin2021real}, including mean squared error (MSE), mean absolute difference (MAD), Connectivity (Conn)~\cite{rhemann2009perceptually},  spatial gradient~\cite{rhemann2009perceptually} (Grad). We also incorporate dtSSD \cite{erofeev2015perceptually} to quantify temporal coherence.
For 3D keypoint estimation, we follow \cite{genmo2025,wang2024tram} to report mean per joint position error (MPJPE), procrustes-aligned MPJPE (PA-MPJPE) and acceleration error (Accel) as a proxy metric for motion smoothness.

\subsection{Comparison to the state of the art}
\noindent\textbf{\textit{Surface Normal Estimation.}} 
In Table \ref{tab:normal_eval_all}, our approach compares favorably to existing methods, achieving lower error than prior foundation models specialized in normal estimation for human understanding tasks (Sapiens~\cite{sapienseccv2024}, DAVID~\cite{saleh2025david}). 

\noindent\textbf{\textit{Human Depth Estimation.}}  
{In Table \ref{tab:depth_eval}, our method demonstrates improved performance compared to other foundation models specifically designed for depth estimation 
(MiDas~\cite{ranftl2020towards}, DepthAnythingV2~\cite{yang2024depth}), and those designed specifically for human understanding tasks (Sapiens~\cite{sapienseccv2024}, DAVID~\cite{saleh2025david}).} 

\noindent \textbf{\textit{Soft Foreground Segmentation.}}
As presented in Table~\ref{tab:evalsoftforeground}, our approach achieves SOTA results compared to other methods trained specifically for the foreground segmentation, in both video and image datasets.

\noindent\textbf{\textit{3D Keypoints Prediction.}} 
We evaluated two variants of our model for 3D keypoint prediction: (1) a model built on top of the WAN \cite{wan2025} backbone used in the rest of the experiments, and (2) a model that uses a larger proprietary backbone that is able to process longer videos at a larger spatial resolution\footnote{$192$ frames at the resolution of $1280\times720$.}. Both models achieve competitive performance on standard 3D keypoint estimation benchmarks EMDB~\cite{kaufmann2023emdb} and H3.6M~\cite{h36m_pami}. Specifically, THFM with the WAN backbone achieves PA-MPJPE of 48.7mm and a MPJPE of 71.8mm on EMDB dataset and PA-MPJPE 34.3mm and a MPJPE of 52mm on H3.6M dataset. These results are further improved by switching to a larger backbone with higher resolution. For example, THFM with such larger backbone achieves MPJPE of 68.7mm on the EMDB dataset setting a new state of the art. We report a comprehensive comparison with state-of-the-art for our best variant of \THFM in Table \ref{tab:combined-eval}. Note that the reported results on all datasets are achieved with a single model without training on their corresponding training sets and relying solely on our own synthetic data. We show example qualitative results of 3d keypoint estimation in Figure~\ref{fig:teaser} and Figure~\ref{fig:skiing}.

\begin{table*}[!t]
    \centering
    \begin{minipage}[t]{0.57\textwidth}
        \vspace{0pt} 
        \centering
        \setlength{\tabcolsep}{3pt}
        \resizebox{\textwidth}{!}{
            \begin{tabular}{cl|ccc|ccc}
            \cmidrule[0.75pt]{1-8}
             & & \multicolumn{3}{c}{RICH (24)} & \multicolumn{3}{c}{EMDB (24)}\\
            \cmidrule(lr){3-5} \cmidrule(lr){6-8}
             & Models  & \scriptsize{PA-MPJPE} & \scriptsize{MPJPE}  & \scriptsize{Accel} & \scriptsize{PA-MPJPE} & \scriptsize{MPJPE} & \scriptsize{Accel} \\
            \cmidrule{1-8}
            \multirow{4}{1em}{\rotatebox[origin=c]{90}{per-frame}}
             & CLIFF~\cite{li2022cliff}  & 56.6 & 102.6  & 22.4  & 68.1  & 103.3  & 24.5  \\
             & HybrIK~\cite{li2020hybrik}  & 56.4 & 96.8  & --  & 65.6 & 122.2   & -- \\
             & HMR2.0~\cite{goel2023humans}   & 48.1  & 96.0 & 18.8  & 60.6  & 98.0 & 19.8  \\
             & ReFit~\cite{refit}  & 47.9  & 80.7 & 17.1  & 58.6  & 88.0  & 20.7  \\
            \cmidrule{1-8}
            \multirow{10}{1em}{\rotatebox[origin=c]{90}{temporal}}
             & VIBE~\cite{kocabas2020vibe}  & 68.4  & 120.5 & 21.8  & 81.4  & 125.9 & 26.6  \\
             & TRACE~\cite{sun2023trace}   & --  & --  & -- & 70.9  & 109.9 & 25.5  \\
             & SLAHMR~\cite{ye2023slahmr}   & 52.5  & -- & 9.4 & 69.5  & 93.5 & 7.1\\
             & PACE~\cite{kocabas2024pace} & 49.3     & -- & 8.8 & --    & -- & -- \\
             & WHAM~\cite{shin2024wham}  & 44.3  & 80.0 & 5.3 & 50.4  & 79.7 & 5.3\\
             & GVHMR~\cite{shen2024gvhmr}  & {39.5} & {66.0} & {4.1} & {42.7} & \underline{72.6} &{3.6}        \\
             & TRAM~\cite{wang2024tram}   & -- & -- & -- & 45.7 & 74.4 & 4.9 \\
             & GENMO \cite{genmo2025} & \underline{39.1} & \underline{66.8}& {4.1} & \underline{42.5} & {73.0} & {3.8} \\
             \cmidrule{1-8}
             & \THFM  & \textbf{34.7} & \textbf{55.0} & \textbf{3.1} & 
             \textbf{35.2} & \textbf{54.6} & \textbf{3.4} \\
            \cmidrule[0.75pt]{1-8}
            \end{tabular}
        }
    \end{minipage}
    \hfill
    \begin{minipage}[t]{0.42\textwidth}
        \vspace{0pt} 
        \centering
        \resizebox{\textwidth}{!}{
            \begin{tabular}{l|cc}
            \cmidrule[0.75pt]{1-3}
            & \multicolumn{2}{c}{H36M (14)} \\
            \cmidrule(lr){2-3}
            Models & \scriptsize{PA-MPJPE} & \scriptsize{MPJPE} \\
            \cmidrule{1-3}
            MHFormer~\cite{Li_2022_CVPR} & 34.4 & 43.0 \\
            MixSTE~\cite{zhang2022mixste} & 32.6 & 40.9 \\
            P-STMO~\cite{shan2022p} & 34.4 & 42.8 \\
            STCFormer~\cite{tang20233d} & \underline{31.8} & {40.5} \\
            PoseFormerV2~\cite{zhao2023poseformerv2} & 35.6 & 45.2 \\
            GLA-GCN~\cite{yu2023gla} & 34.8 & 44.4 \\
            MotionBERT~\cite{zhu2023motionbert} & 32.8 & 41.7 \\
            MotionAGFormer-L~\cite{mehraban2024motionagformer} & 32.7 & 41.6 \\
            PoseAnchor~\cite{kim2025poseanchor} & 32.1 & \underline{40.3}\\
            \cmidrule{1-3}
            \THFM & \textbf{27.7} & \textbf{38.3} \\
            \cmidrule[0.75pt]{1-3}
            \end{tabular}
        }
    \end{minipage}

    \caption{We evaluate \textbf{3D keypoints reconstruction} on RICH and EMDB-1 (left) and H3.6M (right) against \textbf{SOTA methods}. Parenthesis denotes body joints used. All errors are in $mm$, except accel ($m/s^2$). For H3.6M testing, please note that 1) modern HPE methods use 2D keypoints as input (CPN \cite{Chen_2018_CVPR} predictions in this case), with no existing pure video methods (that we are aware of) which can obtain competitive results and 2) that our method was the only one not fine-tuned on any real data (including the H3.6M training set).}
    \label{tab:combined-eval}
\end{table*}

\subsection{Ablation Studies}

In this section, we evaluate some of our design choices, including 1) applying the training objective directly on the latent output of the LDM, or in the ambient RGB space output by the decoder; 2) jointly learning multiple tasks at the same time or separately; 3) the choice of the fixed timestep fed into the latent diffusion model. 
We pick the surface normals estimation task for these experiments. 

As shown in Table \ref{tab:normal_eval_all}, applying the loss in the ambient RGB space leads to improved performance. This is because the latent space is a compressed representation where both temporal and spatial information are reduced, making it difficult for supervision to capture subtle and intricate signals. %
Besides, joint learning of multiple tasks results in a slight reduction in performance, although it still outperforms other state-of-the-art methods. In the future, more advanced techniques for handling multiple conflicting objectives~\cite{sener2018multi} could be explored. Moreover, setting the input timestep to the later stages of the diffusion process yields better performance than using those from the beginning, confirming our assumption that a later input timestep better aligns with our use case, where the input latents are noise-free.

\begin{figure}[t!]
  \centering
  \includegraphics[width=0.8\textwidth]{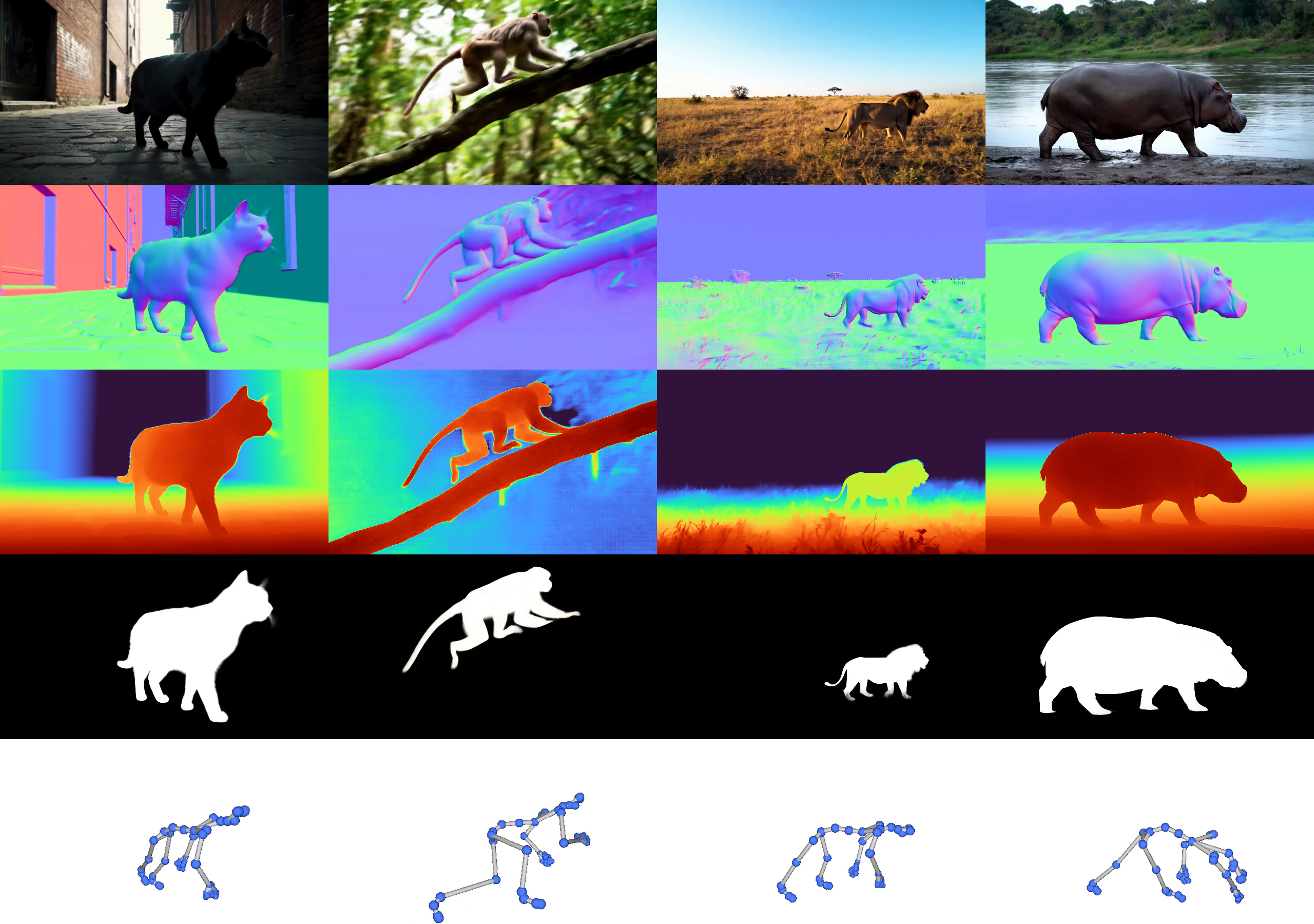}
  \caption{\textbf{Emerging behavior: sim-to-real generalization to OOD classes.} Our approach has been trained only on synthetic videos
    of single class (humans), but generalizes to real-world videos with  a variety of other classes of articulated objects. 
    }
  \label{fig:oodvis}
\end{figure}

\begin{figure}[t!]
  \centering
   \includegraphics[width=0.8\textwidth, trim={0cm 11.75cm 0cm 0cm}, clip]{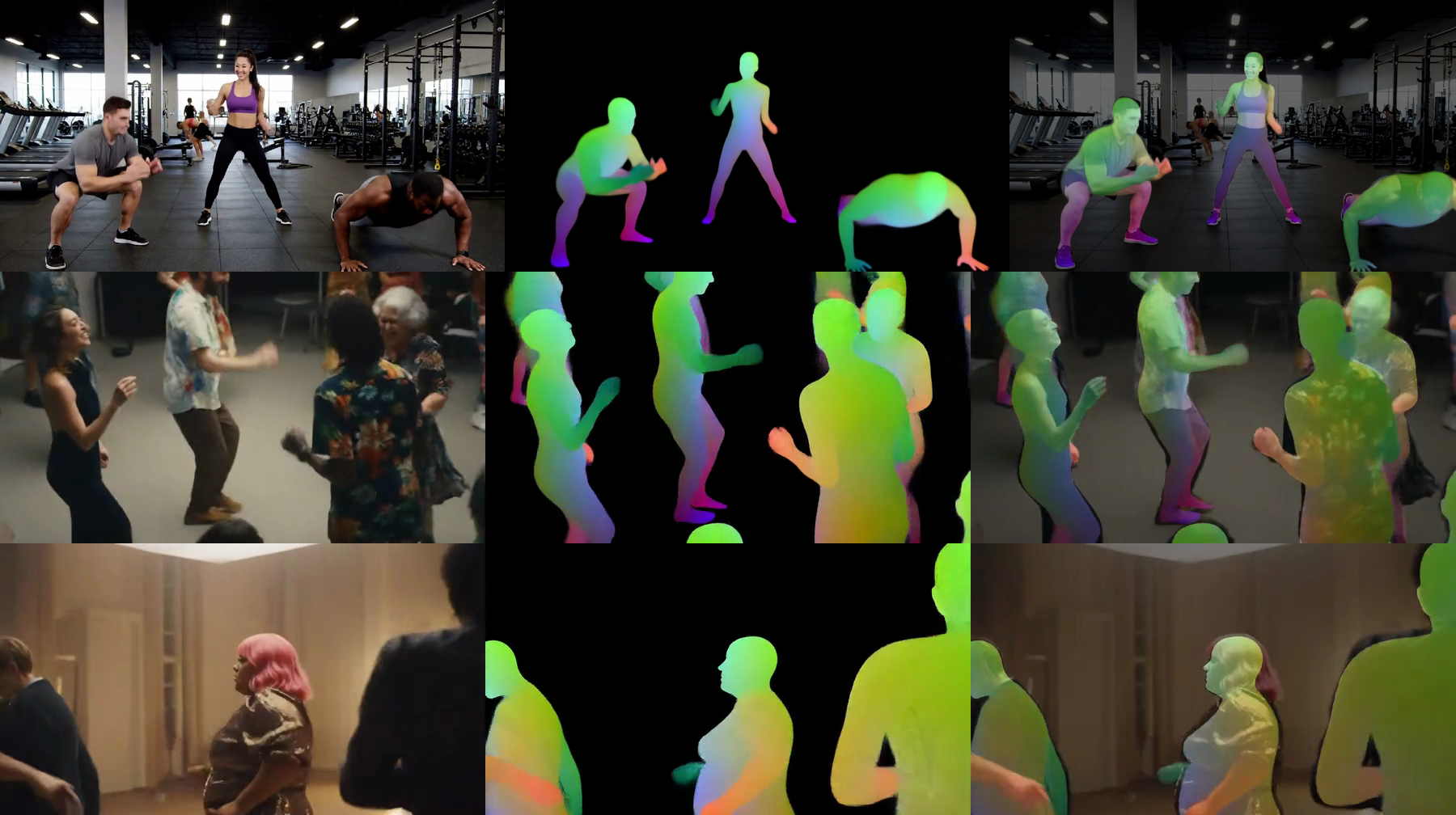}
  \caption{\textbf{Emerging behavior: generalization to multiple instances.} Trained purely on synthetic data with one single human within each video, our method generalizes in zero-shot to real videos with multiple humans. 
  }
  \label{fig:multipeople}
\end{figure}

\subsection{Emerging Behaviors}
In addition to achieving competitive performance that matches or surpasses state-of-the-art methods, our model also exhibits intriguing emergent behaviors beyond its intended training objectives. We believe these behaviors arise from the rich 
representations learned in large-scale text-to-video generation models, shedding light on the broader potential of this paradigm for future exploration.

\noindent\textbf{\textit{Generalization to OOD classes}}:
our approach is trained purely on synthetic videos of human entities, yet it generalizes well to real-world videos—both of humans and of other categories such as animals and anthropomorphic characters—as shown in Figure~\ref{fig:teaser} and Figure~\ref{fig:oodvis}.

\noindent\textbf{\textit{Generalization to multiple instances}}:
trained on synthetic data with one single human within each video, our method generalizes in zero-shot to real videos with multiple humans as shown in Figure~\ref{fig:multipeople}.

\section{Future Work}
Our model achieves competitive performance, for example, matching or surpassing specialized state-of-the-art models when unifying four dense vision tasks; however, we observe reduced performance when unifying dense and sparse vision tasks, and addressing this conflict is a crucial next step that will illuminate the path toward scaling to an even broader range of tasks. Future work will also focus on extending the model to general domains beyond human-centric vision tasks, expanding to cover more diverse tasks, such as language-based video understanding or geometry tasks like camera pose estimation, and exploring the full potential and limits of the video model as a truly unified system.

\section{Conclusions}
In this work, we introduced \THFM, a unified video foundation model that pioneers the use of text-to-video diffusion models as powerful, single-step predictors for multiple perception tasks. We demonstrated that by extending the diffusion architecture and employing learnable tokens alongside text prompting, a single model can simultaneously and effectively address a wide array of both dense (depth, normal, segmentation) and sparse (2d/3d keypoints) vision tasks.

\THFM achieves state-of-the-art performance across human-centric benchmarks, often surpassing specialized models, despite being trained only on synthetic data. \THFM also exhibits emergent generalization capabilities to unseen object categories (like animals and anthropomorphic characters), validating the richness and transferability of video representations learned via diffusion pre-training.

The success of \THFM establishes a clear pathway for developing next-generation video perception foundation models, moving beyond task specialization towards truly generalist architectures. For future work, we plan to investigate techniques for resolving potential conflicts in joint multi-task learning objectives and to further exploit the promptable nature of our model for fine-grained, user-defined interaction and conditional task execution.

\section*{Acknowledgements}
We thank Jeremiah Harmsen, Chris Dyer, João Carreira, Andrew Zisserman, Dima Damen, Thabo Beeler, Di Qiu, Jes\'{u}s P\'{e}rez, Alberto Garc\'{i}a Garc\'{i}a, Sergio Orts Escolano, Erroll Wood, Iker J. de los Mozos, and Emily Conn for their valuable support, feedback, and discussions throughout this project.

\clearpage
{
    \small
    \bibliographystyle{ieeenat_fullname}
    \bibliography{main}
}

\end{document}